\documentclass[10pt,twocolumn,letterpaper]{article}

\usepackage{cvpr}              

\usepackage{graphicx}
\usepackage{amsmath}
\usepackage{amssymb}
\usepackage{booktabs}

\usepackage{multirow,makecell}

\usepackage[table]{xcolor}
\definecolor{mygray}{gray}{0.93}

\usepackage[accsupp]{axessibility}

\usepackage[pagebackref,breaklinks,colorlinks]{hyperref}

\usepackage[capitalize]{cleveref}
\crefname{section}{Sec.}{Secs.}
\Crefname{section}{Section}{Sections}
\Crefname{table}{Table}{Tables}
\crefname{table}{Tab.}{Tabs.}

\def\cvprPaperID{1857} 
\def\confName{CVPR}
\def\confYear{2023}

\begin{document}

\title{Generalized Relation Modeling for Transformer Tracking}

\author{Shenyuan Gao\textsuperscript{1} \quad Chunluan Zhou\textsuperscript{2} \quad Jun Zhang\textsuperscript{1} \\
\textsuperscript{1}The Hong Kong University of Science and Technology \quad \textsuperscript{2}Wormpex AI Research \\
{\tt\small sygao@connect.ust.hk} \quad {\tt\small czhou002@e.ntu.edu.sg} \quad {\tt\small eejzhang@ust.hk}
}

\maketitle

\begin{abstract}
Compared with previous two-stream trackers, the recent one-stream tracking pipeline, which allows earlier interaction between the template and search region, has achieved a remarkable performance gain. However, existing one-stream trackers always let the template interact with all parts inside the search region throughout all the encoder layers. This could potentially lead to target-background confusion when the extracted feature representations are not sufficiently discriminative. To alleviate this issue, we propose a generalized relation modeling method based on adaptive token division. The proposed method is a generalized formulation of attention-based relation modeling for Transformer tracking, which inherits the merits of both previous two-stream and one-stream pipelines whilst enabling more flexible relation modeling by selecting appropriate search tokens to interact with template tokens. An attention masking strategy and the Gumbel-Softmax technique are introduced to facilitate the parallel computation and end-to-end learning of the token division module. Extensive experiments show that our method is superior to the two-stream and one-stream pipelines and achieves state-of-the-art performance on six challenging benchmarks with a real-time running speed. Code and models are publicly available at \href{https://github.com/Little-Podi/GRM}{https://github.com/Little-Podi/GRM}.
\end{abstract}

\begin{figure}[t]
\centering
\includegraphics[width=.93\linewidth]{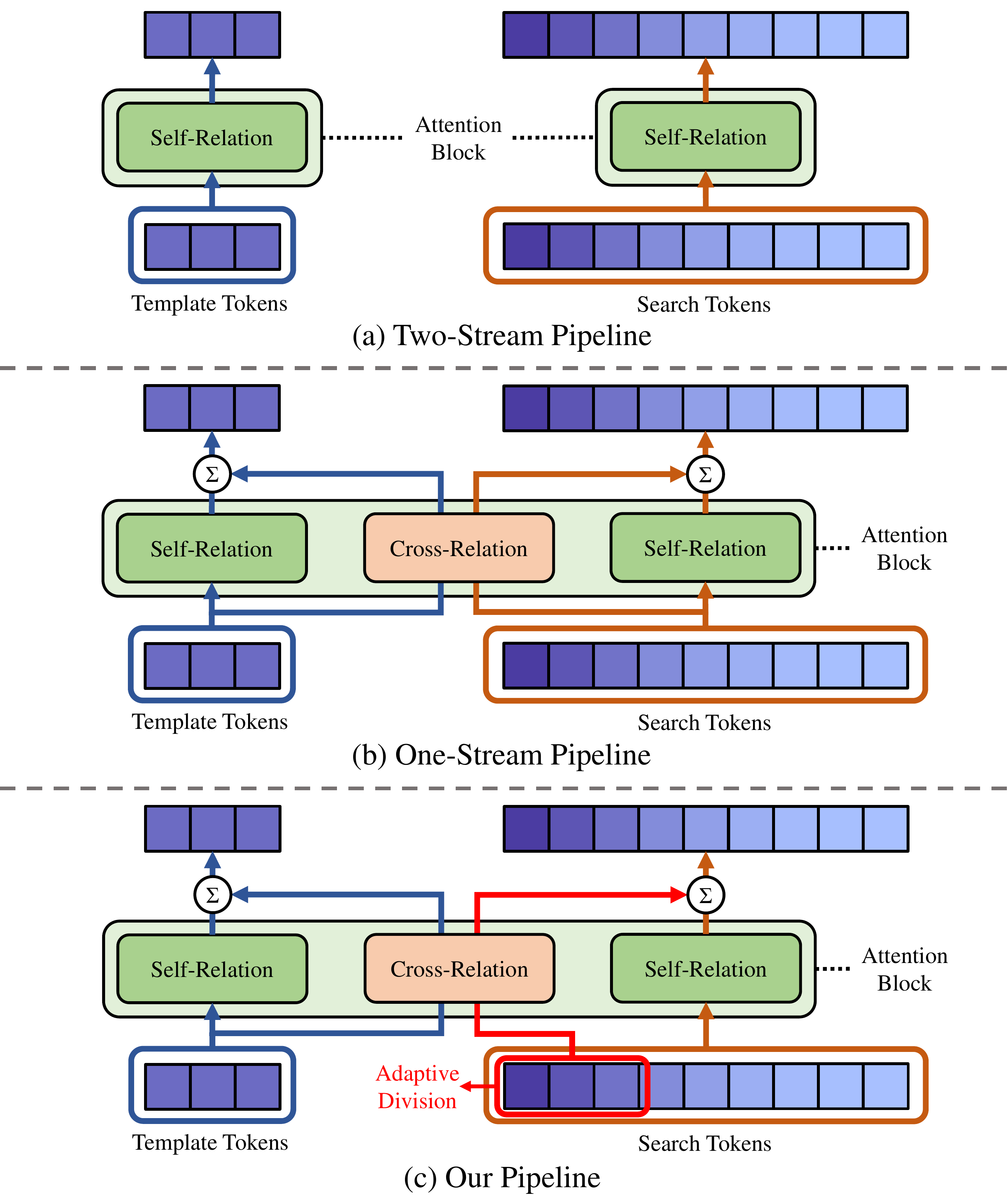}
\vspace{-2.5mm}
\caption{Comparison of different relation modeling pipelines of Transformer-based trackers. The two-stream pipeline uses parallel self-attention blocks to model relations within each set of tokens (template or search tokens). The one-stream pipeline integrates the cross-relation modeling between two sets of tokens and self-relation modeling within each set of tokens via an unified attention block. In contrast, our proposed pipeline performs an adaptive division of the search tokens, which can degenerate to the two-stream form if no search token is selected to interact with template tokens and to the one-stream form if all the search tokens are selected for cross-relation modeling.}
\label{figure-pipeline}
\vspace{-4.5mm}
\end{figure}

\section{Introduction}
Given the target bounding box in the initial frame of a video, visual tracking \cite{javed2022visual} aims to localize the target in successive frames. Over the past few years, two-stream trackers \cite{bertinetto2016fully,li2018high,li2019siamrpn++,zhang2020ocean}, which extract features of the template and search region separately and then model cross-relations of the template and search region in a sequential fashion, have emerged as a dominant tracking paradigm and made a significant progress. Following this two-stream pipeline, several Transformer-based trackers \cite{wang2021transformer,chen2021transformer,gao2022aiatrack} utilize parallel self-attention blocks to enhance the extracted features by modeling global self-relations within each image as illustrated in Fig.~\ref{figure-pipeline}(a). Recently, leveraging the flexibility of the attention mechanism, the one-stream pipeline \cite{cui2022mixformer,ye2022joint,chen2022backbone} is proposed to jointly extract features and model relations, achieving promising performance. By conducting self-attention among all concatenated tokens, both cross-relation modeling and self-relation modeling can be performed simultaneously as illustrated in Fig.~\ref{figure-pipeline}(b).

It is demonstrated in \cite{xie2022correlation,cui2022mixformer,ye2022joint,chen2022backbone} that letting the search region interact with the template as early as possible is beneficial to target-specific feature generation. However, there is no evidence suggesting that all parts inside the search region should always be forced to interact with the template. Actually, due to the cropping strategy \cite{bertinetto2016fully}, there is a large proportion of background inside the search region, where distractors with similar appearance to the target may exist. This would lead to undesired cross-relations between the template and search region as the highly discriminative representations have not been extracted in some early layers.

Although the attention mechanism can inherently weaken improper cross-relations, applying global cross-relation modeling to all layers may still be more or less disruptive. On the one hand, for the search tokens outside the target region, if undesired cross-relations are modeled between the template and distractors, the aggregated features of the distractors may contain the target features from the template, which could cause confusion for precisely identifying the actual target in the search region. On the other hand, for the template tokens, their quality could also be degraded by undesired cross-relations during the iterative update since certain features from the background or even distractors could be aggregated into these tokens. These situations could weaken the target-background discrimination capability of the one-stream pipeline.

Intuitively, only a portion of search tokens, \eg tokens belonging to the target, are suitable for cross-relation modeling when the feature representations are not perfect for target-background discrimination. In some cases, the two-stream relation modeling pipeline could even be better if the feature representations of both the template and search region are imperfect to model cross-relations. The potential limitations of the one-stream pipeline motivates us to ponder: is it really optimal for the template to interact with all parts inside the search region through all encoder layers in the one-stream pipeline?

In this paper, we answer this question by proposing GRM, a generalized relation modeling method that can adaptively select the appropriate search tokens to interact with the template. To be specific, we classify the template and search tokens as three categories. The template tokens form one category while the search tokens are divided into another two categories. Instead of modeling relations within all the tokens as the one-stream pipeline, we restrict the interaction among the three token categories. Only the search tokens that are suitable for cross-relation modeling will interact with the template tokens, whilst the interaction between the remaining search tokens and the template tokens is blocked. With proper divisions, the two-stream pipeline and one-stream pipeline become two degenerated forms of our relation modeling method as discussed in Sec.~\ref{section-contribution1}. Consequently, our method is a generalized formulation of attention-based relation modeling for Transformer tracking, which embraces the advantages of both previous pipelines while being more flexible.

The search token division is performed by a lightweight prediction module, which can adaptively determine which search tokens are suitable for cross-relation modeling based on the input tokens. To accomplish this objective, there are two obstacles to overcome. First, the separate relation modeling for different token categories makes it hard for parallel computation. Second, the discrete token categorization is non-differentiable, thus impeding the end-to-end learning of the token division module. To facilitate parallel computation, we adopt an attention masking strategy to unify the individual attention operations into a single one. Additionally, we introduce the Gumbel-softmax technique \cite{jang2016categorical} to make the discrete token categorization differentiable. Consequently, the search token division module can be implicitly optimized in an end-to-end manner, which promotes its adaptability to deal with different situations.

In summary, our main contributions are three-fold:
\begin{itemize}
    \item[$\bullet$] We present a generalized formulation of relation modeling for Transformer trackers, which divides the input tokens into three categories and enables more flexible interaction between the template and search region.
    \item[$\bullet$] To realize the generalized relation modeling, we devise a token division module to adaptively classify the input tokens. An attention masking strategy and the Gumbel-Softmax technique are introduced to facilitate the parallel computation and end-to-end learning of the proposed module.
    \item[$\bullet$] We conduct extensive experiments and analyses to validate the efficacy of our method. The proposed GRM exhibits outstanding results on six challenging visual tracking benchmarks.
\end{itemize}

\section{Related Works}

\subsection{Visual Tracking Paradigms}
Over the past few years, Siamese trackers \cite{bertinetto2016fully,li2018high,danelljan2019atom} have gained much popularity. Typically, they adopt a two-stream pipeline to separately extract the features of the template and search region. Cross-relations between the two streams are modeled by additional correlation modules. To exploit the power of highly discriminative features, most Siamese trackers \cite{zhang2019deeper,li2019siamrpn++,bhat2019learning,zhang2020ocean} use deep neural networks as the backbone, \eg ResNet-50 \cite{he2016deep}. Recently, the Transformer \cite{vaswani2017attention} architecture has achieved promising results in visual tracking and has become the de-facto choice for many high-performance trackers \cite{wang2021transformer,chen2021transformer,yan2021learning,mayer2022transforming,gao2022aiatrack,yan2022towards,lin2021swintrack}.

To further enhance the feature interaction, several attempts \cite{yu2020deformable,xie2022correlation,guo2022learning} have investigated cross-relation modeling inside the backbone. Recently, another thread of progress \cite{yan2021learning,lin2021swintrack} concatenates the template and search tokens to conduct cross-relation modeling and self-relation modeling jointly. Inspired by these explorations, more recent trackers \cite{cui2022mixformer,ye2022joint,chen2022backbone} adopt a one-stream pipeline to jointly extract the features and model the relations of both the template and search region by the self-attention mechanism. Based on this pipeline, they can utilize advanced pretrained models, \eg MAE \cite{he2022masked}, instead of randomly initialized correlation modules for cross-relation modeling, thereby yielding a remarkable performance gain. Akin to most previous two-stream methods, however, these one-stream methods also treat the search region as a whole, which means that the template always interacts with all parts inside the search region. This may unfavorably cause target-background confusion when the feature representations are imperfect for cross-relation modeling. To mitigate such confusion, we devise a lightweight module to predict which search tokens are appropriate for the cross-relation modeling with the template in each encoder layer. We also show that both the two-stream and one-stream pipelines become degenerated forms of our method.

\subsection{Trackers with Dynamic Designs}
There are several trackers leveraging dynamic designs at different stages for various purposes. Among them, LightTrack \cite{yan2021lighttrack} and AutoMatch \cite{zhang2021learn} use neural architecture search techniques during the training stage. Particularly, LightTrack aims to find a more efficient structure of the network backbone and prediction head, whilst AutoMatch aims to automatically design a more powerful correlation module. Both the network architectures of these trackers are frozen after the training stage. On the contrary, our method introduces a dynamic architecture design that can conduct flexible adaptations during the inference stage.

Benefiting from the flexibility of the attention mechanism, several efforts \cite{liu2021swin,rao2021dynamicvit,yu2022boat,kong2021spvit} have been made to design different token organization methods to accelerate the inference speed of vision Transformer. To make it well-suited to the task of visual tracking, SparseTT \cite{fu2022sparsett} and OSTrack \cite{ye2022joint} utilize the attention weights that are free to borrow from its self-attention calculation. Concretely, SparseTT only aggregates the features from the most similar tokens to improve the discrimination ability, whilst OSTrack eliminates the background tokens to boost efficiency. However, these operations do not have learnable parameters and need to be hand-crafted. In addition, a large proportion of search tokens are still consistently involved in the cross-relation modeling with the template, which may lead to suboptimal feature aggregation since undesired interaction are implicated when the feature representations are not sufficiently discriminative. Differently, we propose an end-to-end learnable token division for Transformer visual tracking, which can adaptively select appropriate search tokens for cross-relation modeling with the template.

\section{Method}

\subsection{Preliminary}
We first revisit the recent one-stream pipeline \cite{cui2022mixformer,ye2022joint,chen2022backbone}, which achieves remarkable tracking performance. The input of the one-stream pipeline is a pair of images, namely template $\boldsymbol{z} \in \mathbb{R}^{H_{z} \times W_{z} \times 3}$ and search region $\boldsymbol{x} \in \mathbb{R}^{H_{x} \times W_{x} \times 3}$. They are first divided into $N_{z}$ and $N_{x}$ non-overlapping image patches of resolution $P \times P$ respectively, where $N_{z} = H_{z}W_{z} / P^{2}$ and $N_{x} = H_{x}W_{x} / P^{2}$. Then, a linear projection is applied to these image patches to generate template patch embeddings $\mathbf{E}_{z} \in \mathbb{R}^{N_{z} \times C}$ and search patch embeddings $\mathbf{E}_{x} \in \mathbb{R}^{N_{x} \times C}$, where $C$ is the embedding dimension. Next, two learnable position embeddings are added to $\mathbf{E}_{z}$ and $\mathbf{E}_{x}$ respectively to embed the spatial information. With some abuse of notation, we still refer to the two updated embeddings as $\mathbf{E}_{z}$ and $\mathbf{E}_{x}$, which are the initial token embeddings of the template and search region, respectively. Afterward, all these tokens are concatenated as a sequence with a length of $N_{z} + N_{x}$ and fed to an encoder. Each encoder layer updates the input tokens via a multi-head attention (MHA) block and a feed-forward network (FFN). Formally, the operations of the $l$-th encoder layer can be expressed as:
\begin{equation}
\begin{aligned}
    \boldsymbol{q} & = \boldsymbol{k} = \boldsymbol{v} = [\mathbf{E}^{l}_{z}; \mathbf{E}^{l}_{x}], \\
    [\mathbf{E'}^{l}_{z}; \mathbf{E'}^{l}_{x}] & = [\mathbf{E}^{l}_{z}; \mathbf{E}^{l}_{x}] + \text{MHA}(\boldsymbol{q}, \boldsymbol{k}, \boldsymbol{v}), \\
    [\mathbf{E}^{l + 1}_{z}; \mathbf{E}^{l + 1}_{x}] & = [\mathbf{E'}^{l}_{z}; \mathbf{E'}^{l}_{x}] + \text{FFN}([\mathbf{E'}^{l}_{z}; \mathbf{E'}^{l}_{x}]), \\
\end{aligned}
\end{equation}
where $\mathbf{E}^{l}_{z}$ and $\mathbf{E}^{l}_{x}$ are the input tokens of the $l$-th encoder layer, and $[;]$ denotes the concatenation operation. We use $\boldsymbol{q}$, $\boldsymbol{k}$ and $\boldsymbol{v}$ to represent the queries, keys and values fed to the multi-head attention block. Since the template and search tokens are jointly processed by the multi-head attention block, the cross-relation and self-relation modeling are seamlessly integrated in each encoder layer. Finally, the output search tokens from the last encoder layer are decoupled from the template tokens and re-shaped to a 2D feature map according to their original spatial positions. The feature map is then taken as the input of to a convolutional head for target bounding box prediction.

\begin{figure}[t]
\centering
\includegraphics[width=.93\linewidth]{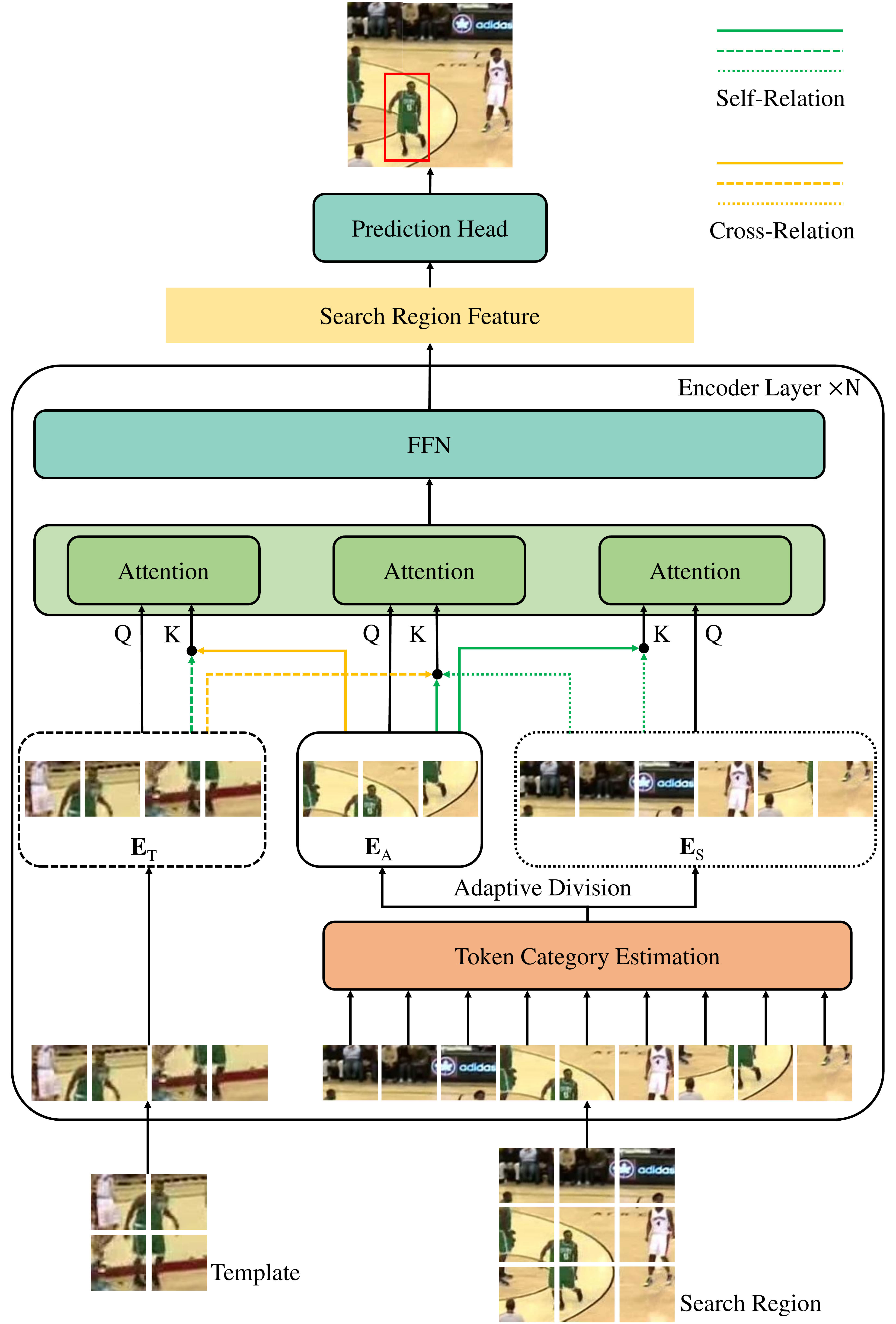}
\vspace{-2.5mm}
\caption{Tracking framework with our generalized relation modeling method. The attention operations for the three token categories are simultaneously conducted with the proposed attention masking strategy. We only illustrate the queries and keys for the attention operations of different token categories. The values are identical to the corresponding keys and are omitted for clarity.}
\label{figure-framework}
\vspace{-4.5mm}
\end{figure}

\subsection{Generalized Relation Modeling}
\label{section-contribution1}
The one-stream pipeline allows free interaction between template tokens and search tokens via the attention mechanism in every encoder layer, which facilitates target-specific feature extraction. However, the large proportion of background inside the search region could lead to undesired cross-relations as the highly discriminative representations have not been extracted in some early layers. Although the attention mechanism can inherently weaken improper cross-relations, they may still cause adverse effects. On the one hand, a certain amount of information from the template is aggregated into the search region that belong to background or distractors, which could increase the difficulty of identifying the target. On the other hand, the representation of the template tokens could be disctracted by the information from the inappropriate search tokens, which degrades the quality of the template during the iterative update. Therefore, it may not be optimal for the template to always interact with all parts inside the search region in every encoder layer.

To mitigate the aforementioned issues, a straightforward solution is to divide the input tokens $\mathbf{E}$ into two categories, denoted by $\mathbf{E}_{T}$ and $\mathbf{E}_{S}$. The first category $\mathbf{E}_{T}$ contains all the template tokens and a portion of search tokens for which the relation modeling with the template tokens are necessary, whilst the second category $\mathbf{E}_{S}$ contains the remaining search tokens. Then, the attention mechanism can be applied to model relations within tokens of each category. This token division can help prevent the undesired interaction between the template tokens and inappropriate search tokens. However, if we directly apply this binary division, there is a risk that the relation modeling between some relevant search tokens would be blocked. For example, when the search tokens of one object are separated into two sets, the interaction between these two sets would not be allowed, which is adverse to feature aggregation. 

To overcome the downside of this binary division, we introduce another token category, denoted by $\mathbf{E}_{A}$, to ensure sufficient interaction between relevant search tokens while preventing the undesired interaction between template tokens and inappropriate search tokens. We divide the search tokens into two sets, one is assigned to category $\mathbf{E}_{A}$ and the other is assigned to category $\mathbf{E}_{S}$. We fix all the template tokens as the category $\mathbf{E}_{T}$ (\ie $\mathbf{E}_{T} = \mathbf{E}_{z}$). The relation modeling rules for the three categories are defined as:
\begin{enumerate}
    \item For category $\mathbf{E}_{T}$, its tokens can aggregate information from the tokens in $\mathbf{E}_{T}$ and $\mathbf{E}_{A}$:
\begin{equation}
\begin{aligned}
    \boldsymbol{q} & = \mathbf{E}^{l}_{T} = \mathbf{E}^{l}_{z}, \\
    \boldsymbol{k} & = \boldsymbol{v} = [\mathbf{E}^{l}_{T}; \mathbf{E}^{l}_{A}] = [\mathbf{E}^{l}_{z}; \mathbf{E}^{l}_{A}], \\
    \mathbf{E'}^{l}_{T} & = \mathbf{E}^{l}_{T} + \text{MHA}(\boldsymbol{q}, \boldsymbol{k}, \boldsymbol{v}).
\end{aligned}
\label{equation-foreground}
\end{equation}
    \item For category $\mathbf{E}_{S}$, its tokens can aggregate information from the tokens in $\mathbf{E}_{S}$ and $\mathbf{E}_{A}$:
\begin{equation}
\begin{aligned}
    \boldsymbol{q} & = \mathbf{E}^{l}_{S}, \\
    \boldsymbol{k} & = \boldsymbol{v} = [\mathbf{E}^{l}_{S}; \mathbf{E}^{l}_{A}] = \mathbf{E}^{l}_{x}, \\
    \mathbf{E'}^{l}_{S} & = \mathbf{E}^{l}_{S} + \text{MHA}(\boldsymbol{q}, \boldsymbol{k}, \boldsymbol{v}).
\end{aligned}
\label{equation-background}
\end{equation}
    \item For category $\mathbf{E}_{A}$, its tokens can aggregate information from all tokens:
\begin{equation}
\begin{aligned}
    \boldsymbol{q} & = \mathbf{E}^{l}_{A}, \\
    \boldsymbol{k} & = \boldsymbol{v} = [\mathbf{E}^{l}_{T}; \mathbf{E}^{l}_{S}; \mathbf{E}^{l}_{A}] = [\mathbf{E}^{l}_{z}; \mathbf{E}^{l}_{x}], \\
    \mathbf{E'}^{l}_{A} & = \mathbf{E}^{l}_{A} + \text{MHA}(\boldsymbol{q}, \boldsymbol{k}, \boldsymbol{v}).
\end{aligned}
\label{equation-shared}
\end{equation}
\end{enumerate}

With $\mathbf{E'}^{l}_{T}$, $\mathbf{E'}^{l}_{S}$ and $\mathbf{E'}^{l}_{A}$, the output tokens of the $l$-th encoder are obtained by:
\begin{equation}
\begin{aligned}
    \mathbf{E'}^{l}_{z} & = \mathbf{E'}^{l}_{T}, \\
    \mathbf{E'}^{l}_{x} & = [\mathbf{E'}^{l}_{S}; \mathbf{E'}^{l}_{A}], \\
    [\mathbf{E}^{l + 1}_{z}; \mathbf{E}^{l + 1}_{x}] & = [\mathbf{E'}^{l}_{z}; \mathbf{E'}^{l}_{x}] + \text{FFN}([\mathbf{E'}^{l}_{z}; \mathbf{E'}^{l}_{x}]).
\end{aligned}
\end{equation}

Fig.~\ref{figure-framework} illustrates our generalized relation modeling method for the three token categories in each encoder layer. The template tokens in $\mathbf{E}_{T}$ and search tokens in $\mathbf{E}_{A}$ can interact with each other via the attention operations defined in Eq.~\eqref{equation-foreground} and Eq.~\eqref{equation-shared}. The interaction between the template tokens and the remaining search tokens in $\mathbf{E}_{S}$ are blocked. All the search tokens can fully interact with each other via the attention operations defined in Eq.~\eqref{equation-background} and Eq.~\eqref{equation-shared}.

Our proposed relation modeling method is a generalized formulation of previous two-stream and one-stream pipelines. Concretely, for the two-stream form, our method degenerates to it when all the search tokens are assigned to $\mathbf{E}_{S}$. The template tokens interact with themselves via an attention operation (Eq.~\eqref{equation-foreground}) and the search tokens interact with themselves via another attention operation (Eq.~\eqref{equation-background}), forming two-stream relation modeling pipeline. As the category $\mathbf{E}_{A}$ is empty, no cross-relation modeling is performed. For the one-stream form, our method degenerates to it when all the search tokens are assigned to $\mathbf{E}_{A}$. In this case, the category $\mathbf{E}_{S}$ contains no search tokens and the operations in Eq.~\eqref{equation-foreground} and Eq.~\eqref{equation-shared} can be combined as a single attention operation. Thus, all the template tokens and search tokens can interact with each other, forming one-stream relation modeling pipeline. Our generalized design can provide higher flexibility to handle different situations in different layers. As shown in Sec.~\ref{section-visualization}, our relation modeling method can adaptively degenerate to the two-stream and one-stream forms when necessary.

\subsection{Adaptive Token Division}
\label{section-contribution2}
To realize our generalized relation modeling method, we need a way to dynamically divide search tokens into either category $\mathbf{E}_{S}$ or category $\mathbf{E}_{A}$. To achieve this, we devise a learnable prediction module for each encoder layer. In particular, to provide target-related cues for search tokens, we generate a target-aware representation by aggregating all the template tokens with global maximum pooling. The target-aware representation is then concatenated with each search token. Afterward, we send the concatenation to a light-weight multi-layer perceptron (MLP) for predicting the probabilities of search tokens belonging to category $\mathbf{E}_{S}$ and category $\mathbf{E}_{A}$:
\begin{equation}
    \pi = \text{Softmax}(\text{MLP}([\text{MaxPool}(\mathbf{E}_{z}); \mathbf{E}_{x}])) \in \mathbb{R}^{N_{x} \times 2},
\label{equation-estimation}
\end{equation}
where $\pi_{i,0}$ and $\pi_{i,1}$ denote the probabilities of the $i$-th search token being assigned to $\mathbf{E}_{S}$ and $\mathbf{E}_{A}$, respectively. With the prediction $\pi$, we determine the category of the $i$-th search token by selecting the category with a higher probability.

Although it seems to be straightforward to apply the prediction module to perform search token division, it is non-trivial for implementation because there are still two obstacles to tackle. That are, how to parallelize the computation for GPU platforms and how to learn the proposed prediction module through end-to-end optimization.

The first obstacle arises from the different relation modeling rules for the three token categories. Generally, the numbers of queries and keys are different for the three categories, making it hard to parallelize the three attention operations in Eq.~\eqref{equation-foreground}, Eq.~\eqref{equation-background} and Eq.~\eqref{equation-shared}. In practice, we empirically find that the speed will be significantly slowed down if we conduct the attention operations for the three categories separately. Inspired by DynamicViT \cite{rao2021dynamicvit}, we adopt an attention masking strategy to tackle this problem. We first translate probabilities $\pi$ to the one-hot categorization $\mathbf{D} \in \{0, 1\}^{N_{x} \times 2}$. Then, we define two one-hot tensors, $\mathbf{D}_{z} \in \{0, 1\}^{N_{z} \times 3}$ and $\mathbf{D}_{x} \in \{0, 1\}^{N_{x} \times 3}$, which represent the categorization of template tokens and that of search tokens, respectively. The first column of $\mathbf{D}_{z}$ is filled with ones and the other two columns are set to zeros, indicating all the template tokens are fixed as $\mathbf{E}_{T}$. The one-hot tensor $\mathbf{D}_{x}$ is obtained by appending a column of zeros to the front of the binary categorization $\mathbf{D}$, meaning no search tokens will be classified as $\mathbf{E}_{T}$. With the categorization of all the tokens $\widehat{\mathbf{D}}=[\mathbf{D}_{z};\mathbf{D}_{x}] \in \{0, 1\}^{(N_{z} + N_{x}) \times 3}$, we then construct an attention mask $\mathbf{M} \in \{0, 1\}^{(N_{z} + N_{x}) \times (N_{z} + N_{x})}$, where each value ${M}_{i,j}$ is calculated by:
\begin{equation}
\begin{aligned}
    \mathbf{M}_{i,j} = & \widehat{\mathbf{D}}_{i, 0}(\widehat{\mathbf{D}}_{j, 0} + \widehat{\mathbf{D}}_{j, 2}) + \widehat{\mathbf{D}}_{i, 1}(\widehat{\mathbf{D}}_{j, 1} + \widehat{\mathbf{D}}_{j, 2}) \\
    & + \widehat{\mathbf{D}}_{i, 2}(\widehat{\mathbf{D}}_{j, 0} + \widehat{\mathbf{D}}_{j, 1} + \widehat{\mathbf{D}}_{j, 2}).
\end{aligned}
\end{equation}

It is easy to verify that $\mathbf{M}_{i,j}$ indicates whether token $i$ can aggregate information from token $j$ according to the relation modeling rules in Eq.~\eqref{equation-foreground}, Eq.~\eqref{equation-background} and Eq.~\eqref{equation-shared}. By computing the Hadamard products of the attention mask and the attention weight matrix of all input tokens, we can combine the three attention operations into a single one with the same functionality. The introduced attention masking strategy can facilitate computation parallelization on GPUs with a minor increase in computational cost.

Second, the discrete categorization of search tokens is non-differentiable, which impedes the optimization of the prediction module. To overcome this drawback, we apply the Gumbel-Softmax technique \cite{jang2016categorical} to draw samples from a categorical distribution with class probabilities $\pi$ by a reparameterization trick:
\begin{equation}
    \mathbf{D} = \text{One-hot}(\mathop{\arg\max}_{i}[g_{i} + \operatorname{log}\pi_{i}]) \in \{0, 1\}^{N_{x} \times 2},
\label{equation-argmax}
\end{equation}
where $g \sim \text{Gumbel}(0, 1)$ and the argmax operation is replaced by the softmax operation as a continuous and differentiable approximation:
\begin{equation}
    \widehat{y}_{i} = \frac{\text{exp}((\operatorname{log}(\pi_{i}) + g_{i}) / \tau)}{\sum^{N_{x}}_{j}\text{exp}((\operatorname{log}(\pi_{j}) + g_{j}) / \tau)}.
\label{equation-softmax}
\end{equation}

During the training stage, the discrete categorization $\mathbf{D}$ sampled by Eq.~\eqref{equation-argmax} is used to divide search tokens in the forward pass and gradients are computed from the continuous Gumbel-Softmax approximation in Eq.~\eqref{equation-softmax} in the backward pass. We refer readers to \cite{jang2016categorical} for the detailed derivation. Thus, the division of search tokens can be implicitly learned by the supervision from the target localization loss presented in Sec.~\ref{section-head} in an end-to-end manner.

\subsection{Target Prediction Head}
\label{section-head}
Our target prediction head is adopted from \cite{ye2022joint}, which receives a 2D feature map to predict the location of the target. It consists of three convolutional branches which are responsible for center classification, offset regression and size regression, respectively. The center classification branch outputs a score map, where each score represents the confidence of the target center locating at the corresponding position. The offset regression branch is utilized to compensate for the discretization error. The size regression branch predicts the height and width of the target. The position with the highest confidence in the center score map is selected as the target position and the corresponding regressed coordinates are used to compute a bounding box as the final prediction.

During the training stage, the classification branch is supervised by a Gaussian map generated according to the center of the ground truth. For the two regression branches, only the coordinates selected to obtain the final prediction are taken account in the regression loss. The total loss of the prediction head is a weighted sum of a focal loss \cite{lin2017focal} for bounding box center classification, a GIoU loss \cite{rezatofighi2019generalized} and a L1 loss for bounding box coordinate regression:
\begin{equation}
    L_{total} = \lambda_{center}L_{focal} + \lambda_{giou}L_{giou} + \lambda_{l1}L_{l1},
\end{equation}
where $\lambda_{center}$, $\lambda_{giou}$ and $\lambda_{l1}$ are trade-off weights to balance the joint optimization.

\begin{table*}[t]
\footnotesize
\centering
\caption{State-of-the-art comparison on GOT-10k, TrackingNet, LaSOT and AVisT. The best two results are shown in \textcolor{red}{red} and \textcolor{blue}{blue} fonts, respectively. We use * to denote that the results on GOT-10k are obtained following the official one-shot protocol.}
\vspace{-2.5mm}
\begin{tabular}{c|c|ccc|ccc|ccc|ccc}
\hline
\rowcolor{mygray}
& & \multicolumn{3}{c|}{GOT-10k* \cite{huang2019got}}
& \multicolumn{3}{c|}{TrackingNet \cite{muller2018trackingnet}}
& \multicolumn{3}{c|}{LaSOT \cite{fan2019lasot}}
& \multicolumn{3}{c}{AVisT \cite{noman2022avist}} \\
\rowcolor{mygray}
\multirow{-2}{*}{Tracker}
& \multirow{-2}{*}{Source}
& AO & SR$_{0.5}$ & SR$_{0.75}$
& AUC & P$_{\text{Norm}}$ & P
& AUC & P$_{\text{Norm}}$ & P
& AUC & OP$50$ & OP$75$ \\
\hline
GRM & Ours & \textcolor{red}{73.4} & \textcolor{red}{82.9} & \textcolor{red}{70.4} & \textcolor{red}{84.0} & \textcolor{red}{88.7} & \textcolor{red}{83.3} & \textcolor{red}{69.9} & \textcolor{blue}{79.3} & \textcolor{red}{75.8} & \textcolor{red}{54.5} & \textcolor{red}{63.1} & \textcolor{red}{45.2} \\
OSTrack \cite{ye2022joint} & ECCV'22 & \textcolor{blue}{71.0} & \textcolor{blue}{80.4} & \textcolor{blue}{68.2} & \textcolor{blue}{83.1} & 87.8 & 82.0 & 69.1 & 78.7 & \textcolor{blue}{75.2} & - & - & - \\
AiATrack \cite{gao2022aiatrack} & ECCV'22 & 69.6 & 80.0 & 63.2 & 82.7 & 87.8 & 80.4 & 69.0 & \textcolor{red}{79.4} & 73.8 & - & - & - \\
SimTrack \cite{chen2022backbone} & ECCV'22 & 68.6 & 78.9 & 62.4 & 82.3 & 86.5 & - & \textcolor{blue}{69.3} & 78.5 & 74.0 & - & - & - \\
RTS \cite{paul2022robust} & ECCV'22 & - & - & - & 81.6 & 86.0 & 79.4 & 69.7 & 76.2 & 73.7 & 50.8 & 55.7 & 38.9 \\
Unicorn \cite{yan2022towards} & ECCV'22 & - & - & - & 83.0 & 86.4 & \textcolor{blue}{82.2} & 68.5 & 76.6 & 74.1 & - & - & - \\
MixFormer \cite{cui2022mixformer} & CVPR'22 & 70.7 & 80.0 & 67.8 & \textcolor{blue}{83.1} & \textcolor{blue}{88.1} & 81.6 & 69.2 & 78.7 & 74.7 & \textcolor{blue}{53.7} & \textcolor{blue}{63.0} & \textcolor{blue}{43.0} \\
ToMP \cite{mayer2022transforming} & CVPR'22 & - & - & - & 81.2 & 86.2 & 78.6 & 67.6 & 78.0 & 72.2 & 51.6 & 59.5 & 38.9 \\
SBT \cite{xie2022correlation} & CVPR'22 & 69.9 & 80.4 & 63.6 & - & - & - & 65.9 & - & 70.0 & - & - & - \\
CSWinTT \cite{song2022transformer} & CVPR'22 & 69.4 & 78.9 & 65.4 & 81.9 & 86.7 & 79.5 & 66.2 & 75.2 & 70.9 & - & - & - \\
STARK \cite{yan2021learning} & ICCV'21 & 68.0 & 77.7 & 62.3 & 81.3 & 86.1 & 78.1 & 66.4 & 76.3 & 71.2 & 51.1 & 59.2 & 39.1 \\
KeepTrack \cite{mayer2021learning} & ICCV'21 & - & - & - & - & - & - & 67.1 & 77.2 & 70.2 & 49.4 & 56.3 & 37.8 \\
AutoMatch \cite{zhang2021learn} & ICCV'21 & 65.2 & 76.6 & 54.3 & 76.0 & - & 72.6 & 58.3 & - & 59.9 & - & - & - \\
TransT \cite{chen2021transformer} & CVPR'21 & 67.1 & 76.8 & 60.9 & 81.4 & 86.7 & 80.3 & 64.9 & 73.8 & 69.0 & 49.0 & 56.4 & 37.2 \\
Alpha-Refine \cite{yan2021alpha} & CVPR'21 & - & - & - & 80.5 & 85.6 & 78.3 & 65.3 & 73.2 & 68.0 & 49.6 & 55.7 & 38.2 \\
TMT \cite{wang2021transformer} & CVPR'21 & 67.1 & 77.7 & 58.3 & 78.4 & 83.3 & 73.1 & 63.9 & - & 61.4 & 48.1 & 55.3 & 33.8 \\
Ocean \cite{zhang2020ocean} & ECCV'20 & 61.1 & 72.1 & 47.3 & - & - & - & 56.0 & 65.1 & 56.6 & 38.9 & 43.6 & 20.5 \\
PrDiMP \cite{danelljan2020probabilistic} & CVPR'20 & 63.4 & 73.8 & 54.3 & 75.8 & 81.6 & 70.4 & 59.8 & 68.8 & 60.8 & 43.3 & 48.0 & 28.7 \\
SiamAttn \cite{yu2020deformable} & CVPR'20 & - & - & - & 75.2 & 81.7 & - & 56.0 & 64.8 & - & - & - & - \\
DiMP \cite{bhat2019learning} & ICCV'19 & 61.1 & 71.7 & 49.2 & 74.0 & 80.1 & 68.7 & 56.9 & 65.0 & 56.7 & 41.9 & 45.7 & 26.0 \\
ATOM \cite{danelljan2019atom} & CVPR'19 & - & - & - & 70.3 & 77.1 & 64.8 & 51.5 & 57.6 & 50.5 & 38.6 & 41.5 & 22.2 \\
SiamRPN++ \cite{li2019siamrpn++} & CVPR'19 & 51.7 & 61.6 & 32.5 & 73.3 & 80.0 & 69.4 & 49.6 & 56.9 & 49.1 & 39.0 & 43.5 & 21.2 \\
\hline
\end{tabular}
\label{table-main}
\vspace{-2.5mm}
\end{table*}

\begin{table*}[t]
\footnotesize
\centering
\caption{Comparison with the state-of-the-art trackers on NfS30 and UAV123 in terms of AUC score.}
\vspace{-2.5mm}
\begin{tabular}{c|cccccccc|c}
\hline
\rowcolor{mygray}
Tracker & SiamRPN++ \cite{li2019siamrpn++} & ATOM \cite{danelljan2019atom} & DiMP \cite{bhat2019learning} & TransT \cite{chen2021transformer} & TMT \cite{wang2021transformer} & STARK \cite{yan2021learning} & ToMP \cite{mayer2022transforming} & OSTrack \cite{ye2022joint} & GRM \\
\hline
NfS30 \cite{kiani2017need} & 50.2 & 59.0 & 62.0 & 65.7 & \textcolor{blue}{66.5} & 65.2 & \textcolor{red}{66.9} & 64.7 & 65.6 \\
UAV123 \cite{mueller2016benchmark} & 61.3 & 65.0 & 65.4 & \textcolor{blue}{69.1} & 67.5 & \textcolor{blue}{69.1} & 69.0 & 68.3 & \textcolor{red}{70.2} \\
\hline
\end{tabular}
\label{table-addition}
\vspace{-4.5mm}
\end{table*}

\section{Experiments}

\subsection{Implementation Details}
The architecture of our network backbone is identical to the encoder part of ViT-B \cite{dosovitskiy2020image}. It includes 12 sequential encoder layers, each of which comprises a multi-head attention block and a feed-forward network. We adopt the MAE-pretrained model \cite{he2022masked} to initialize our network backbone for its fast convergence as in \cite{ye2022joint}. The search region is $4^{2}$ times the target object area and resized to a resolution of $256 \times 256$ pixels, whilst the template is $2^{2}$ times the target object area and resized to $128 \times 128$ pixels. The cropped images are then down-sampled by a patch embedding projection with a stride of 16. The feature resolution is the same from the first to the last encoder layer. Each branch of the target prediction head comprises 4 stacked Conv-BN-ReLU layers. We use an identical structure for all token division modules in different encoder layers, which is a simple MLP that includes two hidden layers with GELU activation \cite{hendrycks2016gaussian} and an output layer. The channel dimensions of the hidden layers and output layer are 384, 192 and 2, respectively.

Our experiments are conducted with NVIDIA GeForce RTX 3090 GPUs. In line with most Transformer-based trackers, we use the training splits of LaSOT \cite{fan2019lasot}, TrackingNet \cite{muller2018trackingnet}, GOT-10k \cite{huang2019got}, and COCO \cite{lin2014microsoft} for training except for the GOT-10k evaluation. The whole network is optimized with the AdamW optimizer \cite{loshchilov2017decoupled} for 300 epochs. The initial learning rate is $4 \times 10^{-5}$ for the parameters with the pretrained weights and $4 \times 10^{-4}$ for other randomly initialized parameters. It decays by a factor of 10 after 240 epochs of training. As for GOT-10k, the model is trained for 100 epochs and the learning rate decays at epoch 80.

\subsection{Results and Comparisons}
To show the effectiveness and generalization ability of the proposed method, we evaluate and compare our GRM with several state-of-the-art trackers on six benchmarks. The results are summarized in Table~\ref{table-main}, Table~\ref{table-addition} and Table~\ref{table-performance}.

\noindent\textbf{GOT-10k.} GOT-10k \cite{huang2019got} provides 180 short-term video sequences without publicly available ground truths. To ensure zero overlaps of object classes between training and testing, we strictly follow their one-shot protocol. As shown in Table~\ref{table-main}, our tracker improves all metrics by a large margin. Notably, with the aligned model settings, our tracker outperforms the most competitive tracker OSTrack \cite{ye2022joint} by $2.4\%$ in AO and $2.5\%$ in success rate.

\noindent\textbf{TrackingNet.} TrackingNet \cite{muller2018trackingnet} is a large-scale short-term tracking benchmark. Our results are reported by the online evaluation server. Table~\ref{table-main} shows that our GRM achieves $84.0\%$ in success score and $83.3\%$ in precision score, overtaking all previously published trackers.

\noindent\textbf{LaSOT.} LaSOT \cite{fan2019lasot} is a densely annotated large-scale dataset that contains 280 long-term video sequences for public evaluation. From Table~\ref{table-main}, we find that our method sets a new state-of-the-art on LaSOT, which demonstrates that our approach is also well-suited to the tracking scenarios when the video sequences are extremely long.

\noindent\textbf{AVisT.} AVisT \cite{noman2022avist} is a recently released benchmark, which comprises 120 sequences in diverse scenarios with adverse visibility, such as bad weather conditions and camouflage. Our tracker surpasses the previous best tracker MixFormer \cite{cui2022mixformer}, which evidences that our method has a stronger capability in handling these challenging scenarios.

\noindent\textbf{NfS30.} Need for Speed (NfS) \cite{kiani2017need} comprises 100 videos with fast-moving objects. As previous works, we evaluate our method on its low frame rate version NfS30. The results in Table~\ref{table-addition} show that our method achieves comparable performance with the state-of-the-art trackers, demonstrating the generalization ability of GRM.

\noindent\textbf{UAV123.} UAV123 \cite{mueller2016benchmark} contains 123 video sequences captured from a low-altitude unmanned aerial vehicle perspective. It is also a long-term tracking dataset with an average sequence length of 915 frames. The result of our method is shown in Table~\ref{table-addition}. The top-ranked performance shows the strong ability of our tracker to perform long-term tracking.

\noindent\textbf{Scalability.} To show our scalability to larger architectures and higher resolutions, we also implement a variant named GRM-L320. With all other configurations fixed, we change the network backbone to ViT-L \cite{dosovitskiy2020image} and increase the resolution of the search region to $320 \times 320$ pixels with $5^{2}$ times the target object area. Table~\ref{table-performance} illustrates that our variant is highly competitive compared to the performance-oriented variants of most recent works \cite{cui2022mixformer,ye2022joint,chen2022backbone}.

\begin{table}[t]
\scriptsize
\centering
\caption{Comparison with other performance-oriented variants.}
\vspace{-2.5mm}
\begin{tabular}{c|ccc|ccc}
\hline
\rowcolor{mygray}
& \multicolumn{3}{c|}{LaSOT \cite{fan2019lasot}}
& \multicolumn{3}{c}{TrackingNet \cite{muller2018trackingnet}} \\
\rowcolor{mygray}
\multirow{-2}{*}{Tracker}
& AUC & P$_{\text{Norm}}$ & P
& AUC & P$_{\text{Norm}}$ & P \\
\hline
GRM-L320 & \textcolor{red}{71.4} & \textcolor{red}{81.2} & \textcolor{red}{77.9} & \textcolor{red}{84.4} & \textcolor{red}{88.9} & \textcolor{red}{84.0} \\
OSTrack-384 \cite{ye2022joint} & \textcolor{blue}{71.1} & \textcolor{blue}{81.1} & \textcolor{blue}{77.6} & \textcolor{blue}{83.9} & \textcolor{blue}{88.5} & \textcolor{blue}{83.2} \\
SimTrack-L/14 \cite{chen2022backbone} & 70.5 & 79.7 & 76.2 & 83.4 & 87.4 & - \\
MixFormer-L22k \cite{cui2022mixformer} & 70.1 & 79.9 & 76.3 & \textcolor{blue}{83.9} & \textcolor{red}{88.9} & 83.1 \\
\hline
\end{tabular}
\label{table-performance}
\vspace{-4.5mm}
\end{table}

\subsection{Ablation Studies}
To analyze the effectiveness of the proposed method, we conduct detailed ablation studies on GOT-10k \cite{huang2019got}. Our experimental results are summarized in Table~\ref{table-ablation1} and Table~\ref{table-ablation2}.

\noindent\textbf{Search Token Division Strategy.} To demonstrate the superiority of our method over previous two-stream and one-stream pipelines, we first devise two counterparts of them using our definition of token categories (Eq.~\eqref{equation-foreground}, Eq.~\eqref{equation-background} and Eq.~\eqref{equation-shared}). As shown in Table~\ref{table-ablation1}, the two-stream pipeline (variant~\#1) is equivalent to the situation where all the search tokens are consistently assigned to the category $\mathbf{E}_{S}$, and the one-stream pipeline (variant~\#2) is equivalent to the situation where all the search tokens are consistently assigned to the category $\mathbf{E}_{T}$. For a fair comparison, in variant~\#1, we conduct separate attentions for template tokens and search tokens in all encoder layers except for the last one which acts as the correlation module by applying a global attention to all the template and search tokens. Comparing variant~\#1 with variant~\#2, we find that the one-stream pipeline performs much better than the two-stream pipeline with the same number of model parameters. It suggests earlier interaction between the template and search region is essential.

\begin{table}[t]
\footnotesize
\centering
\caption{Ablative experiments of different search token divisions.}
\vspace{-2.5mm}
\begin{tabular}{c|ccc|ccc|c}
\hline
\rowcolor{mygray}
& \multicolumn{3}{c|}{Token Category} & \multicolumn{3}{c|}{GOT-10k* \cite{huang2019got}} & Speed \\
\rowcolor{mygray}
\multirow{-2}{*}{\#} & $\mathbf{E}_{T}$ & $\mathbf{E}_{S}$ & $\mathbf{E}_{A}$ & AO & SR$_{0.5}$ & SR$_{0.75}$ & fps \\
\hline
1 & & \checkmark & & 64.7 & 73.3 & 60.2 & 60 \\
2 & \checkmark & & & 71.0 & 80.3 & 68.2 & 51 \\
\hline
3 & \checkmark & \checkmark & & 71.5 & 80.6 & 68.3 & 45 \\
4 & \checkmark & \checkmark & \checkmark & 71.9 & 81.3 & 69.0 & 44 \\
5 & & \checkmark & \checkmark & \textbf{73.4} & \textbf{82.9} & \textbf{70.4} & 45 \\
\hline
\end{tabular}
\label{table-ablation1}
\vspace{-2.5mm}
\end{table}

\begin{table}[t]
\footnotesize
\centering
\caption{Ablative experiments about where we apply the token division and how we aggregate the target-aware representation.}
\vspace{-2.5mm}
\begin{tabular}{c|c|c|ccc}
\hline
\rowcolor{mygray}
& & & \multicolumn{3}{c}{GOT-10k* \cite{huang2019got}} \\
\rowcolor{mygray}
\multirow{-2}{*}{\#} & \multirow{-2}{*}{Layer} & \multirow{-2}{*}{Aggregation} & AO & SR$_{0.5}$ & SR$_{0.75}$ \\
\hline
a & None & None & 71.0 & 80.3 & 68.2 \\
b & 1 - 12 & Maximum Pooling & 71.2 & 80.6 & 67.8 \\
c & 5 - 12 & Maximum Pooling & 72.3 & 81.6 & 69.3 \\
d & 2 - 12 & Average Pooling & 71.8 & 81.1 & 68.9 \\
e & 2 - 12 & Maximum Pooling & \textbf{73.4} & \textbf{82.9} & \textbf{70.4} \\
\hline
\end{tabular}
\label{table-ablation2}
\vspace{-4.5mm}
\end{table}

\begin{figure}[t]
\centering
\includegraphics[width=.93\linewidth]{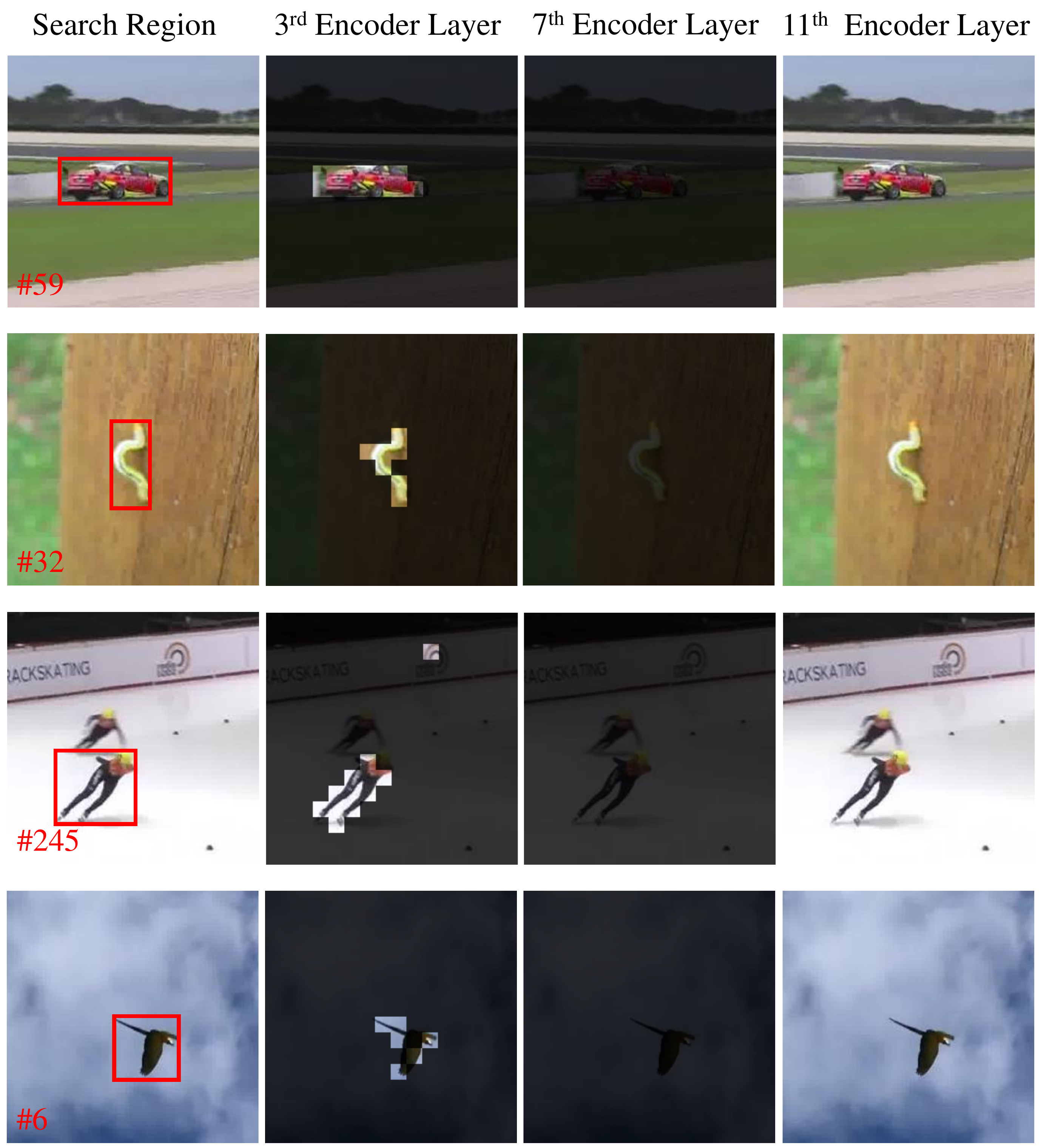}
\vspace{-2.5mm}
\caption{Visualization of some representative search token division results in several encoder layers. We choose the $3^{rd}$, the $7^{th}$ and the $11^{th}$ encoder layer as the examples of the intermediate form, the degenerated two-stream and one-stream forms of our generalized relation modeling. The shaded region denotes the search tokens that are classified as the category $\mathbf{E}_{S}$ and do not interact with the template tokens. The ground-truth bounding boxes and the frame indices are shown in the first figure of each line.}
\label{figure-visualization}
\vspace{-4.5mm}
\end{figure}

We then use the prediction module to classify the search tokens into $\mathbf{E}_{T}$ and $\mathbf{E}_{S}$, which do not interact with each other. Although this division seems to be a possible solution to overcome the limitation of the one-stream pipeline, we find that the improvement of variant~\#3 over variant~\#2 is limited since it would block the interaction of some objects with their relevant context in the search region, impeding the feature extraction. In variant~\#4, the token division module classifies the search tokens into $\mathbf{E}_{T}$, $\mathbf{E}_{S}$ and $\mathbf{E}_{A}$. From the result of variant~\#4, we can find that the limitation of variant~\#3 can be alleviated by introducing the third token category $\mathbf{E}_{A}$. However, it still blocks the interaction between the search tokens in $\mathbf{E}_{T}$ and those in $\mathbf{E}_{S}$. Finally, by dividing search tokens into $\mathbf{E}_{S}$ and $\mathbf{E}_{A}$, the completeness of interaction among the search tokens can be ensured, and the template tokens can still interact with the selected tokens in $\mathbf{E}_{A}$ that are suitable for cross-relation modeling. Thus, the overall performance of variant~\#5 is the best among all variants in Table~\ref{table-ablation1}.

\noindent\textbf{Encoder Layers with Token Division.} To investigate which encoder layers are suitable to conduct adaptive token division, we implement three variants of our method by inserting the prediction module we proposed in Sec.~\ref{section-contribution2} to all encoder layers (variant~\#b), the last eight encoder layers (variant~\#c), and all encoder layers except for the first one (variant~\#e). From Table~\ref{table-ablation2}, we can see that both variant~\#b and variant~\#c are inferior to variant~\#e. The result of variant~\#b suggests that it is inappropriate to perform token division by the prediction module in the first encoder layer where the raw embedding projections have not been refined by any feature aggregation. The result of variant~\#c suggests that our token division method can perform favorably in other earlier encoder layers. Therefore, we conduct adaptive token division in all encoder layers except for the first layer as the default setting in our experiments.

\noindent\textbf{Target-Aware Representation Aggregation.} We also explore the aggregation method to generate target-aware representation, which is used to predict the category of search tokens (Eq.~\eqref{equation-estimation}). From the last two rows of Table~\ref{table-ablation2}, we can see that aggregation via maximum pooling (variant~\#e) is better than average pooling (variant~\#d). The underlying reason might be that the maximum pooling operation is less sensitive to noise, which may be introduced by the surrounding background inside the template image.

\noindent\textbf{Speedup by Attention Masking.} As mentioned in Sec.~\ref{section-contribution2}, separate attention operations will become a bottleneck for the running speed. For a quantitative comparison, we evaluate the inference speed of the two implementations. The speed with separate attention operations is 33 fps, whilst the proposed attention masking strategy accelerates it to 45 fps.

\subsection{Visualizations}
\label{section-visualization}
To investigate how our generalized relation modeling method works in each encoder layer, we visualize some representative search token division results in Fig.~\ref{figure-visualization}. From the visualization results, we observe that: (i) in the $3^{rd}$ encoder layer, our prediction module classifies the search tokens relevant to the target as $\mathbf{E}_{A}$ to perform cross-relation modeling with the template tokens; (ii) in the $7^{th}$ encoder layer, none of the search tokens are selected as $\mathbf{E}_{A}$ and the relation modeling in this layer degenerates to a two-stream form; (iii) in the $11^{th}$ encoder layer, all of the search tokens are selected as $\mathbf{E}_{A}$ and it degenerates to a one-stream form. In fact, we find that the earlier layers tend to perform as the two-stream form or the intermediate form in which a portion of search tokens are selected to interact with the template tokens. Besides, the relation modeling degenerates to the one-stream form in the last five encoder layers. This suggests that highly discriminative representations have been extracted in the deeper layers, and the constraints on global interaction are not necessary for these layers. More thorough discussions can be found in the appendices.

\section{Conclusion}
In this paper, we present a generalized relation modeling method, which inherits the strengths of both the two-stream and one-stream Transformer tracking pipelines whilst being more flexible. An end-to-end optimized token division module with attention masking strategy is introduced to perform adaptive token division. Extensive experiments and analyses demonstrate the efficacy of the proposed method. We hope this work would shed some new light on promoting the relation modeling for Transformer trackers.

{\small
\bibliographystyle{ieee_fullname}
\bibliography{egbib}

\begin{thebibliography}{10}\itemsep=-1pt

\bibitem{bertinetto2016fully}
Luca Bertinetto, Jack Valmadre, Joao~F Henriques, Andrea Vedaldi, and Philip~HS
  Torr.
\newblock Fully-convolutional siamese networks for object tracking.
\newblock In {\em European conference on computer vision}, pages 850--865.
  Springer, 2016.

\bibitem{bhat2019learning}
Goutam Bhat, Martin Danelljan, Luc~Van Gool, and Radu Timofte.
\newblock Learning discriminative model prediction for tracking.
\newblock In {\em Proceedings of the IEEE/CVF international conference on
  computer vision}, pages 6182--6191, 2019.

\bibitem{chen2022backbone}
Boyu Chen, Peixia Li, Lei Bai, Lei Qiao, Qiuhong Shen, Bo Li, Weihao Gan, Wei
  Wu, and Wanli Ouyang.
\newblock Backbone is all your need: A simplified architecture for visual
  object tracking.
\newblock {\em arXiv preprint arXiv:2203.05328}, 2022.

\bibitem{chen2021transformer}
Xin Chen, Bin Yan, Jiawen Zhu, Dong Wang, Xiaoyun Yang, and Huchuan Lu.
\newblock Transformer tracking.
\newblock In {\em Proceedings of the IEEE/CVF Conference on Computer Vision and
  Pattern Recognition}, pages 8126--8135, 2021.

\bibitem{cui2022mixformer}
Yutao Cui, Cheng Jiang, Limin Wang, and Gangshan Wu.
\newblock Mixformer: End-to-end tracking with iterative mixed attention.
\newblock In {\em Proceedings of the IEEE/CVF Conference on Computer Vision and
  Pattern Recognition}, pages 13608--13618, 2022.

\bibitem{danelljan2019atom}
Martin Danelljan, Goutam Bhat, Fahad~Shahbaz Khan, and Michael Felsberg.
\newblock Atom: Accurate tracking by overlap maximization.
\newblock In {\em Proceedings of the IEEE/CVF Conference on Computer Vision and
  Pattern Recognition}, pages 4660--4669, 2019.

\bibitem{danelljan2020probabilistic}
Martin Danelljan, Luc~Van Gool, and Radu Timofte.
\newblock Probabilistic regression for visual tracking.
\newblock In {\em Proceedings of the IEEE/CVF conference on computer vision and
  pattern recognition}, pages 7183--7192, 2020.

\bibitem{dosovitskiy2020image}
Alexey Dosovitskiy, Lucas Beyer, Alexander Kolesnikov, Dirk Weissenborn,
  Xiaohua Zhai, Thomas Unterthiner, Mostafa Dehghani, Matthias Minderer, Georg
  Heigold, Sylvain Gelly, et~al.
\newblock An image is worth 16x16 words: Transformers for image recognition at
  scale.
\newblock {\em arXiv preprint arXiv:2010.11929}, 2020.

\bibitem{fan2019lasot}
Heng Fan, Liting Lin, Fan Yang, Peng Chu, Ge Deng, Sijia Yu, Hexin Bai, Yong
  Xu, Chunyuan Liao, and Haibin Ling.
\newblock Lasot: A high-quality benchmark for large-scale single object
  tracking.
\newblock In {\em Proceedings of the IEEE/CVF conference on computer vision and
  pattern recognition}, pages 5374--5383, 2019.

\bibitem{fu2022sparsett}
Zhihong Fu, Zehua Fu, Qingjie Liu, Wenrui Cai, and Yunhong Wang.
\newblock Sparsett: Visual tracking with sparse transformers.
\newblock {\em arXiv preprint arXiv:2205.03776}, 2022.

\bibitem{gao2022aiatrack}
Shenyuan Gao, Chunluan Zhou, Chao Ma, Xinggang Wang, and Junsong Yuan.
\newblock Aiatrack: Attention in attention for transformer visual tracking.
\newblock In {\em European Conference on Computer Vision}, pages 146--164.
  Springer, 2022.

\bibitem{guo2022learning}
Mingzhe Guo, Zhipeng Zhang, Heng Fan, Liping Jing, Yilin Lyu, Bing Li, and
  Weiming Hu.
\newblock Learning target-aware representation for visual tracking via
  informative interactions.
\newblock {\em arXiv preprint arXiv:2201.02526}, 2022.

\bibitem{he2022masked}
Kaiming He, Xinlei Chen, Saining Xie, Yanghao Li, Piotr Doll{\'a}r, and Ross
  Girshick.
\newblock Masked autoencoders are scalable vision learners.
\newblock In {\em Proceedings of the IEEE/CVF Conference on Computer Vision and
  Pattern Recognition}, pages 16000--16009, 2022.

\bibitem{he2016deep}
Kaiming He, Xiangyu Zhang, Shaoqing Ren, and Jian Sun.
\newblock Deep residual learning for image recognition.
\newblock In {\em Proceedings of the IEEE conference on computer vision and
  pattern recognition}, pages 770--778, 2016.

\bibitem{hendrycks2016gaussian}
Dan Hendrycks and Kevin Gimpel.
\newblock Gaussian error linear units (gelus).
\newblock {\em arXiv preprint arXiv:1606.08415}, 2016.

\bibitem{huang2019got}
Lianghua Huang, Xin Zhao, and Kaiqi Huang.
\newblock Got-10k: A large high-diversity benchmark for generic object tracking
  in the wild.
\newblock {\em IEEE Transactions on Pattern Analysis and Machine Intelligence},
  43(5):1562--1577, 2019.

\bibitem{jang2016categorical}
Eric Jang, Shixiang Gu, and Ben Poole.
\newblock Categorical reparameterization with gumbel-softmax.
\newblock {\em arXiv preprint arXiv:1611.01144}, 2016.

\bibitem{javed2022visual}
Sajid Javed, Martin Danelljan, Fahad~Shahbaz Khan, Muhammad~Haris Khan, Michael
  Felsberg, and Jiri Matas.
\newblock Visual object tracking with discriminative filters and siamese
  networks: a survey and outlook.
\newblock {\em IEEE Transactions on Pattern Analysis and Machine Intelligence},
  2022.

\bibitem{kiani2017need}
Hamed Kiani~Galoogahi, Ashton Fagg, Chen Huang, Deva Ramanan, and Simon Lucey.
\newblock Need for speed: A benchmark for higher frame rate object tracking.
\newblock In {\em Proceedings of the IEEE International Conference on Computer
  Vision}, pages 1125--1134, 2017.

\bibitem{kong2021spvit}
Zhenglun Kong, Peiyan Dong, Xiaolong Ma, Xin Meng, Wei Niu, Mengshu Sun, Bin
  Ren, Minghai Qin, Hao Tang, and Yanzhi Wang.
\newblock Spvit: Enabling faster vision transformers via soft token pruning.
\newblock {\em arXiv preprint arXiv:2112.13890}, 2021.

\bibitem{li2019siamrpn++}
Bo Li, Wei Wu, Qiang Wang, Fangyi Zhang, Junliang Xing, and Junjie Yan.
\newblock Siamrpn++: Evolution of siamese visual tracking with very deep
  networks.
\newblock In {\em Proceedings of the IEEE/CVF Conference on Computer Vision and
  Pattern Recognition}, pages 4282--4291, 2019.

\bibitem{li2018high}
Bo Li, Junjie Yan, Wei Wu, Zheng Zhu, and Xiaolin Hu.
\newblock High performance visual tracking with siamese region proposal
  network.
\newblock In {\em Proceedings of the IEEE conference on computer vision and
  pattern recognition}, pages 8971--8980, 2018.

\bibitem{lin2021swintrack}
Liting Lin, Heng Fan, Yong Xu, and Haibin Ling.
\newblock Swintrack: A simple and strong baseline for transformer tracking.
\newblock {\em arXiv preprint arXiv:2112.00995}, 2021.

\bibitem{lin2017focal}
Tsung-Yi Lin, Priya Goyal, Ross Girshick, Kaiming He, and Piotr Doll{\'a}r.
\newblock Focal loss for dense object detection.
\newblock In {\em Proceedings of the IEEE international conference on computer
  vision}, pages 2980--2988, 2017.

\bibitem{lin2014microsoft}
Tsung-Yi Lin, Michael Maire, Serge Belongie, James Hays, Pietro Perona, Deva
  Ramanan, Piotr Doll{\'a}r, and C~Lawrence Zitnick.
\newblock Microsoft coco: Common objects in context.
\newblock In {\em European conference on computer vision}, pages 740--755.
  Springer, 2014.

\bibitem{liu2021swin}
Ze Liu, Yutong Lin, Yue Cao, Han Hu, Yixuan Wei, Zheng Zhang, Stephen Lin, and
  Baining Guo.
\newblock Swin transformer: Hierarchical vision transformer using shifted
  windows.
\newblock In {\em Proceedings of the IEEE/CVF International Conference on
  Computer Vision}, pages 10012--10022, 2021.

\bibitem{loshchilov2017decoupled}
Ilya Loshchilov and Frank Hutter.
\newblock Decoupled weight decay regularization.
\newblock {\em arXiv preprint arXiv:1711.05101}, 2017.

\bibitem{mayer2022transforming}
Christoph Mayer, Martin Danelljan, Goutam Bhat, Matthieu Paul, Danda~Pani
  Paudel, Fisher Yu, and Luc Van~Gool.
\newblock Transforming model prediction for tracking.
\newblock In {\em Proceedings of the IEEE/CVF Conference on Computer Vision and
  Pattern Recognition}, pages 8731--8740, 2022.

\bibitem{mayer2021learning}
Christoph Mayer, Martin Danelljan, Danda~Pani Paudel, and Luc Van~Gool.
\newblock Learning target candidate association to keep track of what not to
  track.
\newblock In {\em Proceedings of the IEEE/CVF International Conference on
  Computer Vision}, pages 13444--13454, 2021.

\bibitem{mueller2016benchmark}
Matthias Mueller, Neil Smith, and Bernard Ghanem.
\newblock A benchmark and simulator for uav tracking.
\newblock In {\em European conference on computer vision}, pages 445--461.
  Springer, 2016.

\bibitem{muller2018trackingnet}
Matthias Muller, Adel Bibi, Silvio Giancola, Salman Alsubaihi, and Bernard
  Ghanem.
\newblock Trackingnet: A large-scale dataset and benchmark for object tracking
  in the wild.
\newblock In {\em Proceedings of the European conference on computer vision
  (ECCV)}, pages 300--317, 2018.

\bibitem{noman2022avist}
Mubashir Noman, Wafa~Al Ghallabi, Daniya Najiha, Christoph Mayer, Akshay
  Dudhane, Martin Danelljan, Hisham Cholakkal, Salman Khan, Luc Van~Gool, and
  Fahad~Shahbaz Khan.
\newblock Avist: A benchmark for visual object tracking in adverse visibility.
\newblock {\em arXiv preprint arXiv:2208.06888}, 2022.

\bibitem{paul2022robust}
Matthieu Paul, Martin Danelljan, Christoph Mayer, and Luc Van~Gool.
\newblock Robust visual tracking by segmentation.
\newblock In {\em Computer Vision--ECCV 2022: 17th European Conference, Tel
  Aviv, Israel, October 23--27, 2022, Proceedings, Part XXII}, pages 571--588.
  Springer, 2022.

\bibitem{rao2021dynamicvit}
Yongming Rao, Wenliang Zhao, Benlin Liu, Jiwen Lu, Jie Zhou, and Cho-Jui Hsieh.
\newblock Dynamicvit: Efficient vision transformers with dynamic token
  sparsification.
\newblock {\em Advances in neural information processing systems},
  34:13937--13949, 2021.

\bibitem{rezatofighi2019generalized}
Hamid Rezatofighi, Nathan Tsoi, JunYoung Gwak, Amir Sadeghian, Ian Reid, and
  Silvio Savarese.
\newblock Generalized intersection over union: A metric and a loss for bounding
  box regression.
\newblock In {\em Proceedings of the IEEE/CVF conference on computer vision and
  pattern recognition}, pages 658--666, 2019.

\bibitem{song2022transformer}
Zikai Song, Junqing Yu, Yi-Ping~Phoebe Chen, and Wei Yang.
\newblock Transformer tracking with cyclic shifting window attention.
\newblock In {\em Proceedings of the IEEE/CVF Conference on Computer Vision and
  Pattern Recognition}, pages 8791--8800, 2022.

\bibitem{vaswani2017attention}
Ashish Vaswani, Noam Shazeer, Niki Parmar, Jakob Uszkoreit, Llion Jones,
  Aidan~N Gomez, {\L}ukasz Kaiser, and Illia Polosukhin.
\newblock Attention is all you need.
\newblock {\em Advances in neural information processing systems}, 30, 2017.

\bibitem{wang2021transformer}
Ning Wang, Wengang Zhou, Jie Wang, and Houqiang Li.
\newblock Transformer meets tracker: Exploiting temporal context for robust
  visual tracking.
\newblock In {\em Proceedings of the IEEE/CVF Conference on Computer Vision and
  Pattern Recognition}, pages 1571--1580, 2021.

\bibitem{xie2022correlation}
Fei Xie, Chunyu Wang, Guangting Wang, Yue Cao, Wankou Yang, and Wenjun Zeng.
\newblock Correlation-aware deep tracking.
\newblock In {\em Proceedings of the IEEE/CVF Conference on Computer Vision and
  Pattern Recognition}, pages 8751--8760, 2022.

\bibitem{yan2022towards}
Bin Yan, Yi Jiang, Peize Sun, Dong Wang, Zehuan Yuan, Ping Luo, and Huchuan Lu.
\newblock Towards grand unification of object tracking.
\newblock {\em arXiv preprint arXiv:2207.07078}, 2022.

\bibitem{yan2021learning}
Bin Yan, Houwen Peng, Jianlong Fu, Dong Wang, and Huchuan Lu.
\newblock Learning spatio-temporal transformer for visual tracking.
\newblock In {\em Proceedings of the IEEE/CVF International Conference on
  Computer Vision}, pages 10448--10457, 2021.

\bibitem{yan2021lighttrack}
Bin Yan, Houwen Peng, Kan Wu, Dong Wang, Jianlong Fu, and Huchuan Lu.
\newblock Lighttrack: Finding lightweight neural networks for object tracking
  via one-shot architecture search.
\newblock In {\em Proceedings of the IEEE/CVF Conference on Computer Vision and
  Pattern Recognition}, pages 15180--15189, 2021.

\bibitem{yan2021alpha}
Bin Yan, Xinyu Zhang, Dong Wang, Huchuan Lu, and Xiaoyun Yang.
\newblock Alpha-refine: Boosting tracking performance by precise bounding box
  estimation.
\newblock In {\em Proceedings of the IEEE/CVF Conference on Computer Vision and
  Pattern Recognition}, pages 5289--5298, 2021.

\bibitem{ye2022joint}
Botao Ye, Hong Chang, Bingpeng Ma, Shiguang Shan, and Xilin Chen.
\newblock Joint feature learning and relation modeling for tracking: A
  one-stream framework.
\newblock In {\em European Conference on Computer Vision}, pages 341--357.
  Springer, 2022.

\bibitem{yu2022boat}
Tan Yu, Gangming Zhao, Ping Li, and Yizhou Yu.
\newblock Boat: Bilateral local attention vision transformer.
\newblock {\em arXiv preprint arXiv:2201.13027}, 2022.

\bibitem{yu2020deformable}
Yuechen Yu, Yilei Xiong, Weilin Huang, and Matthew~R Scott.
\newblock Deformable siamese attention networks for visual object tracking.
\newblock In {\em Proceedings of the IEEE/CVF conference on computer vision and
  pattern recognition}, pages 6728--6737, 2020.

\bibitem{zhang2022not}
Yifan Zhang, Qingyong Hu, Guoquan Xu, Yanxin Ma, Jianwei Wan, and Yulan Guo.
\newblock Not all points are equal: Learning highly efficient point-based
  detectors for 3d lidar point clouds.
\newblock In {\em Proceedings of the IEEE/CVF Conference on Computer Vision and
  Pattern Recognition}, pages 18953--18962, 2022.

\bibitem{zhang2021learn}
Zhipeng Zhang, Yihao Liu, Xiao Wang, Bing Li, and Weiming Hu.
\newblock Learn to match: Automatic matching network design for visual
  tracking.
\newblock In {\em Proceedings of the IEEE/CVF International Conference on
  Computer Vision}, pages 13339--13348, 2021.

\bibitem{zhang2019deeper}
Zhipeng Zhang and Houwen Peng.
\newblock Deeper and wider siamese networks for real-time visual tracking.
\newblock In {\em Proceedings of the IEEE/CVF Conference on Computer Vision and
  Pattern Recognition}, pages 4591--4600, 2019.

\bibitem{zhang2020ocean}
Zhipeng Zhang, Houwen Peng, Jianlong Fu, Bing Li, and Weiming Hu.
\newblock Ocean: Object-aware anchor-free tracking.
\newblock In {\em European Conference on Computer Vision}, pages 771--787.
  Springer, 2020.

\end{thebibliography}
}

\clearpage

\appendix
\noindent\textbf{\LARGE Appendix} \\
\renewcommand{\thesection}{\Alph{section}}

For total transparency and a better understanding of our work, in this appendix, we supplement discussions about limitations and failure cases, differences with a pioneering work, as well as potential avenues for further explorations.

\section{Limitations and Failure Cases}
\begin{figure}[htbp]
\centering
\includegraphics[width=.93\linewidth]{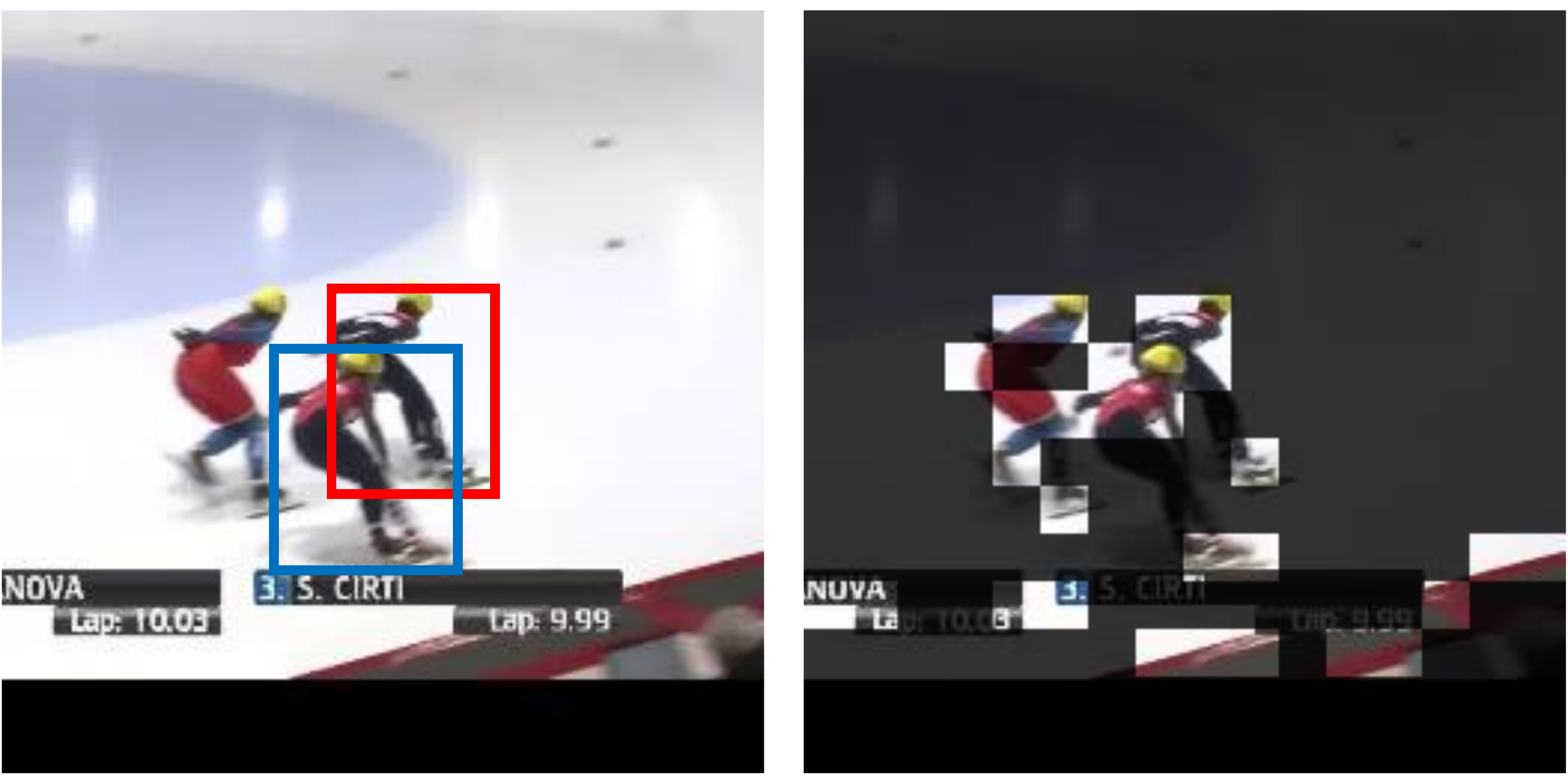}
\vspace{-2.5mm}
\caption{Visualization of a typical failure case.}
\label{figure-failure}
\vspace{-4.5mm}
\end{figure}

Though our adaptive token division is generally helpful, it is still possible that some tokens from distractors are mistakenly selected to interact with the template. This is typically when our tracker might fail. Fig.~\ref{figure-failure} shows a failure case. When the target (\textcolor{red}{red box}) is partially occluded by the distractor (\textcolor{blue}{blue box}), the token selection becomes confusing and our tracker drifts for a while in the subsequent frames. A promising solution is using model update to exploit useful temporal cues, which we leave for future work.

\section{Differences with DynamicViT}
One may wonder about the differences between DynamicViT \cite{rao2021dynamicvit} and our GRM since the former method also incorporates an attention masking strategy and the Gumbel-Softmax technique. It is worth noting that the proposed method is driven by a completely different motivation. In particular, contrary to DynamicViT, which is designed to accelerate the inference speed with a definite accuracy drop, our method aims to prevent undesired cross-relations and thus improves the tracking performance. With this totally different objective, our method differs from DynamicViT in three major aspects: First, besides the task-specific (image classification) loss, DynamicViT needs three extra losses to constrain the token sparsification, whereas our method can implicitly learn the adaptive token division solely by the task-specific (target localization) loss. Second, in DynamicViT, once a token is pruned in a certain layer, it will never be used in the subsequent layers. Differently, our token division is independently determined by each layer, making it more robust to possible improper token selection in the earlier layers as it can still be recovered in the latter layers. Third, during inference, DynamicViT abandons a fixed ratio of tokens regardless of the prediction scores, whereas our method classifies an adaptive portion of tokens into each category based on the predictions.

\section{Possible Explorations}

\subsection{Supervision of Token Division}
Honestly, the outset of this work mainly stems from IA-SSD \cite{zhang2022not}, which selects the sparse yet important foreground points for efficient 3D object detection. As a prototype in our early explorations, the prediction modules in each encoder layer are directly supervised by the ground-truth bounding boxes and do not affect the global cross-relation modeling during training. Essentially, they are learning a rough foreground classification based on each feature vector. During inference, we sort all the search tokens according to their foreground classification scores. Then, a portion of search tokens with higher scores is selected to model cross-relations with the template tokens. The remaining search tokens thus form another token group and model self-relations with themselves. Inspired by BOAT \cite{yu2022boat}, we also maintain some shared search tokens around the selection boundary to allow the information flow between these two search token groups. Although the actual foreground classification results are far from precise, we surprisingly find a performance gain. We thus argue that selecting partial search tokens could be helpful as long as we refer to a relatively reasonable token ranking (\eg foreground classification scores). Nonetheless, the performance of this approach seems to be unstable on different benchmarks and the ratios of token selection in every encoder layer need to be tuned with lots of manual efforts.

Taking these explorations as foundations, the proposed GRM resorts to the Gumbel-Softmax technique to enable end-to-end optimized token division and achieves satisfactory results on multiple benchmarks. Nevertheless, the supervision from the ground-truth bounding boxes can be simultaneously applied to our token division modules as an auxiliary guidance, which is not investigated in this work. We conjecture that the integration of both implicit and explicit supervisions could lead to more interpretable token division results and contribute to better tracking performance.

\subsection{Form of Relation Modeling}
From the converged model, we notice that the relation modeling in most encoder layers permanently degenerates to two-stream or one-stream forms. For those degenerated cases, the learning of token division actually becomes a neural architecture search process. The calculation of the corresponding token division modules can be skipped during the inference to further reduce the computation overhead and reach a faster running speed than we reported. It is also noteworthy that there appears to be no obvious pattern in the form of relational modeling from the earlier layers to the latter layers. The irregular pattern may depend upon the initialization weights from the pretrained models.

We also attempt to use some regularizations (\eg the entropy of the token division ratio) to encourage all layers behave as the intermediate form. However, the eventual results become sensitive and we cannot witness on-par performance across all benchmarks. The underlying reason might be that for most encoder layers, the two degenerated forms can generalize better since each learned attention block only needs to handle a stationary situation, namely either global cross-relation modeling or no cross-relation modeling.

\subsection{Alternative to Discrete Categorization}
Prompted by one of the reviewers, we realize that we can also use the continuous estimations of the prediction modules to scale the raw attention weights. This flexible alternative can naturally bypass the non-differentiable obstacle caused by the strict constraints in our method and eliminate the need of Gumbel-Softmax. In contrast to the token-level attention weights, the scaling weights here can be devised at region-level to serve as a better complement. Such hierarchical attention weight design may hold more promises to improve the relation modeling for Transformer trackers, which is worth investigating in future work.

\end{document}